\documentclass[preprint,12pt,5p,twocolumns]{elsarticle}

\usepackage{amssymb}
\usepackage{prletters}
\usepackage{glossaries}
\usepackage{float}
\usepackage[moderate,tracking=normal]{savetrees}
\newacronym{gan}{GAN}{Generative Adversarial Network}
\newacronym{mia}{MIA}{Membership Inference Attack}
\newacronym{ulmia}{IIA}{Identity Inference Attack}
\newacronym{ml}{ML}{Machine Learning}
\newacronym{cgan}{C-GAN}{Convolutional Generative Adversarial Network}
\newacronym{is}{IS}{Inception Score}
\newacronym{dl}{DL}{Deep Learning}

\journal{}

\newcommand{\sm}[1]{{\leavevmode\color{black} #1}}
\newcommand{\smbis}[1]{{\leavevmode\color{black} #1}}

\newcommand{\dm}[1]{{\leavevmode\color{black} #1}}

\begin{document}

\begin{frontmatter}

\title{Fingerprint Membership and Identity Inference Against \\ Generative Adversarial Networks\tnoteref{pnrr}}
\tnotetext[pnrr]{The work was partially supported by the European Union under the Italian National Recovery and Resilience Plan (NRRP) of NextGenerationEU, partnership on “Telecommunications of the Future” (PE00000001 - program “RESTART”).}

\author[math]{Saverio Cavasin}
\ead{saverio.cavasin@phd.unipd.it}

\affiliation[math]{organization={Department of Mathematics, University of Padova},%Department and Organization
            addressline={Via Trieste, 63}, 
            city={Padova},
            postcode={35131}} 

\fntext[cariparo]{Daniele Mari's activities were supported by Fondazione CaRiPaRo under the grants “Dottorati di Ricerca” 2021/2022.}
            
\author[dei,cariparo]{Daniele Mari}
\ead{daniele.mari@dei.unipd.it}

\author[dei]{Simone Milani}
\ead{simone.milani@dei.unipd.it}

\affiliation[dei]{organization={Department of Information Engineering, University of Padova},
            addressline={Via Gradenigo 6A}, 
            city={Padova},
            postcode={35131}}

\author[math]{Mauro Conti}
\ead{mauro.conti@unipd.it}

\begin{abstract}

\sm{Generative models are gaining significant attention as potential catalysts for a novel industrial revolution.
 Since automated sample generation can be useful to solve privacy and data scarcity issues that usually affect learned biometric models, such technologies became widely spread in this field.}
\newline
\sm{In this paper, we assess the vulnerabilities of generative machine learning models concerning identity protection by designing and testing an identity inference attack on fingerprint datasets created by means of a generative adversarial network. Experimental results show that the proposed solution proves to be effective under different configurations and easily extendable to other biometric measurements.}

\end{abstract}

\begin{keyword}
Membership Inference \sep Identity Inference \sep Fingerprints \sep Biometrics \sep GANs \sep Media Forensics \sep Explainable forensics

\end{keyword}

\end{frontmatter}

\section{Introduction}
\label{sec:intro}
\smbis{Biometric signals are among the most robust and simple means to perform the identification or the authentication of a user for a wide variety of applications (e.g., home banking, payment apps).}

To increase the accuracy of these systems, \gls{dl} techniques have been explored and introduced also in biometrics \cite{kasar2016face, guo2019survey, liu2020novel, rim2021fingerprint, sundararajan2018deep}, \smbis{but the adoption of learned strategies implies the availability of large datasets (otherwise the final accuracy collapses), as well as the risk of possible information leakage concerning the training data (thus revealing sensitive data about some of the users)}
 \cite{shokri2017membership, choquette2021label, hu2022membership, hayes2017logan}.

\sm{To tackle these problems, multiple researchers have proposed the adoption of generative models, like \glspl{gan} \cite{goodfellow2020generative}, which automatically create new original biometric samples \cite{karras2020analyzing} extending the datasets with new synthetic IDs.  Indeed, scientific literature has recently proposed the use of GANs for faces, voices, fingerprints.} 
\smbis{In this way, biometric training data could be generated by means of some \gls{gan} APIs that provide as many samples as required, thus improving users' privacy since no real acquisitions would need to be shared.}

\smbis{Although this seems to solve the aforementioned problems, the inherent learned nature of such solutions poses the same security threats, since it allows} an attacker to partially identify samples belonging to the original dataset (e.g., \gls{mia} on GANs \cite{hayes2017logan}). 

\sm{The vulnerable nature of biometric data is an issue perfectly addressed by generative methods. By exploiting synthetic data, unnecessary privacy leakage is avoided. For example, in an access control system implemented through fingerprints, the sensors and the software can be trained and tuned with generated samples as long as their properties are representative of the real data they are emulating. Nevertheless, the models that provide such systems still need privacy assessment methodologies in order to protect themself from leaking unwanted information. Indeed, as displayed by our proposed method, the cross comparison of different acquisitions of the same fingers represents a serious threat. Even in a scenario where the subject providing the model is able to protect the original dataset, the attack methodology we carry is still going to be effective, as long as different acquisitions of the same finger are obtained. This means that if the attackers access different samples, by any source, by any mean, they can carry the very same attack against the models trained by a privacy-compliant source. Eventually, such consideration imply that even in the case of a privacy-compliant supply strategy by a given company, their models are still threatened by possible data leakages from different, less compliant companies. }

\sm{In this paper, we push forward this possibility by showing that it is possible to infer the identity of the subjects adopted to train the generative models. This task is relatively new among inference attacks since the attacker does not need to verify if a specific biometric sample has been used to train the algorithm (\gls{mia}): he just needs a new acquisition from a specific user and infer if it could possibly match with the unknown training set (Identity Inference Attack).} 
\sm{This working condition applies to many more realistic scenarios, since the assumption that the attacker possesses some of the training samples (\gls{mia}) is much stronger. Moreover, \gls{ulmia} analysis can be exploited for different purposes that go beyond privacy applications, e.g., highlighting eventual flaws or ill conditions in the training dataset and procedures (overfitting, naive models, poor diversity in the data) \cite{hu2022membership}.}
%Without having access to any of the actual fingerprints in the training dataset, an attacker could find out if a person contributed to its creation.
\sm{We tested such a possibility on fingerprint-generating GANs since inference attacks have not been conducted on this domain before. However, due to the difficulties in acquiring new samples, the use of generative strategies has become increasingly studied thus requiring an analysis of the related inference problem.}

\smbis{The main contributions of the current paper can be summarized as follows.

\

\begin{itemize}
\item \sm{ We propose an \gls{ulmia} attack for biometric generative models that infers whether some unknown biometric samples from a target user were employed in training the analyzed model. 
This work departs from \cite{hayes2017logan} since it is intended to infer the identity of the user, instead of the data membership (i.e., whether a given sample has been used in training). Indeed, \gls{ulmia} proves to be extremely recent and timely \cite{li2022user}, while being at its very first investigation stages. To the best of our knowledge, this is the first work
to tackle such problems on fingerprint GANs.}
\item \sm{ \gls{mia} and \gls{ulmia} have been tested on fingerprint generating \glspl{gan}. This work is the first analysis on information inference  issues for fingerprint generative neural networks, which proves to be crucial given privacy and acquisition difficulties in creating a sufficiently-exhaustive and legally-shareable fingerprint datasets. This analysis is extremely useful with respect to the current state-of-the-art in biometric generative schemes.}
\item \sm{The presented GAN inference attack is optimized using the Inception Score metric thus enabling an attacker to guess the best parameters and strategies with respect to the victim GAN. } 
\end{itemize}}

\

\smbis{In the following, Section~\ref{sec:related} overviews the existing literature on this subject, while Section~\ref{sec:materials} describes the proposed attack. Section~\ref{sec:experimental} introduces the experimental setting and Section~\ref{sec:results} presents the relative results. In the end, Section~\ref{sec:discussion} discusses the possible implications, and Section~\ref{sec:conclusion} draws the final conclusions.}
\section{Related Works}
\label{sec:related}

One of the most recent trends in the privacy protection fields for fingerprint images consists in using \glspl{gan} \cite{goodfellow2020generative} to \smbis{extend the training set for machine learning algorithms.} On the other hand, researchers in the \gls{mia} community have proposed various black and white box attacks on generative models.

\subsection{Generative Algorithms for fingerprints data}
\glspl{gan} have been widely used in the literature as a tool to generate new fingerprints with high-quality minutiae \cite{minaee2018finger,mistry2020fingerprint, sams2022hq,bouzaglo2022synthesis} or to reconstruct partially degraded ones \cite{huang2020latent, joshi2019latent}.
In particular, in \cite{minaee2018finger} they use DC-GAN \cite{radford2015unsupervised} together with a connectivity regularization loss to generate faithful fingerprint images. Then in \cite{mistry2020fingerprint} the authors use I-WGAN \cite{gulrajani2017improved} together with an identity loss that allows to generate fingerprints from the same user. In \cite{sams2022hq}, fingerprints are generated in two steps: first, a StyleGAN-2 \cite{karras2020analyzing} based model is used to produce the skeleton of the fingerprint; then, \cite{zhu2017unpaired} applies CycleGAN to perform style transfer and transform these skeletons into actual images.  
This allows to obtain very high-quality images compared to previous methods.

\smbis{Finally, the approach in  \cite{bouzaglo2022synthesis}  also uses StyleGAN2 to generate samples, but the input signal is created by an encoder that permits selecting the position of the minutiae thus generating fingerprints with consistent identities.}

\subsection{Membership inference attacks}
Given a set of data points, and a machine learning model, \gls{mia} consists in guessing which of the samples were used to train the algorithm i.e., that are members of the training set. In general, the \gls{mia} is said to be white-box if the attacker has access to the weights of the model while it is said to be black-box if he can only analyze the predictions. The latter can be further subdivided by incrementally removing information (e.g., when attacking a model that only returns the top-n predicted classes).

One of the first works to introduce \glspl{mia} is \cite{shokri2017membership} where ``machine learning as a service" classification models were attacked. This is a good example of a black box setting since even the \gls{ml} algorithm is unknown to the attacker. The authors' strategy was to train multiple shadow models (one for each class) to mimic the behavior of the attacked one. Afterward, they train a classifier to assess the membership of the samples using the shadow models as a proxy for the attacked one allowing the classifier to be later used to attack the actual black box cloud API.

Following this idea, many works have proposed more reliable and effective attacks and countermeasures \cite{hu2022membership,yeom2018privacy, salem2018ml}. In \cite{yeom2018privacy} the correctness of the prediction and the loss value are used as discriminative factors, in \cite{salem2018ml} the prediction entropy is used as a feature instead.

Additionally, many other \glspl{mia} have been designed to work on different types of models (e.g., generative ones), for example in \cite{hayes2017logan} the authors propose the first
\gls{mia} attack on \glspl{gan}. In the white-box scenario, the authors use the discriminator prediction confidence as the discriminative factor between members and non-members, while, in the black-box scenario, they train a \gls{gan} to mimic the target model and then they exploit the newly trained discriminator to perform the white box attack. On the other hand in \cite{hilprecht2019monte} the authors use Monte Carlo integration to approximate the probability that the item is a member.

Expanding \glspl{mia} some researchers have shown that it is not only possible to assess the membership of a sample in the training set but also of the user depicted in the photo himself. In particular, in \cite{li2022user} it is shown that it is possible to understand if photos of a user were used to train a metric embedding learned system by looking at how tightly some images of the person are clustered by the network. And additionally, these images don't even need to have been used as training samples, showing that the model is actually memorizing the user identity in its weights. Some additional works have shown that \glspl{ulmia} can be carried out also to assess authorship of samples used to train language models \cite{song2019auditing} and to find out if someone's voice was used to train some voice services \cite{miao2021audio}.
\section{Identity inference on GANs}
\label{sec:materials}

\sm{In order to evaluate the feasibility of identity inference for biometric generative models, we considered \glspl{cgan} approaches since they are the most frequently adopted in biometric applications and allow us to generalize the obtained results to the whole family of models.}

\begin{figure}[!t]
    \centering
    \includegraphics[width=.9\columnwidth]{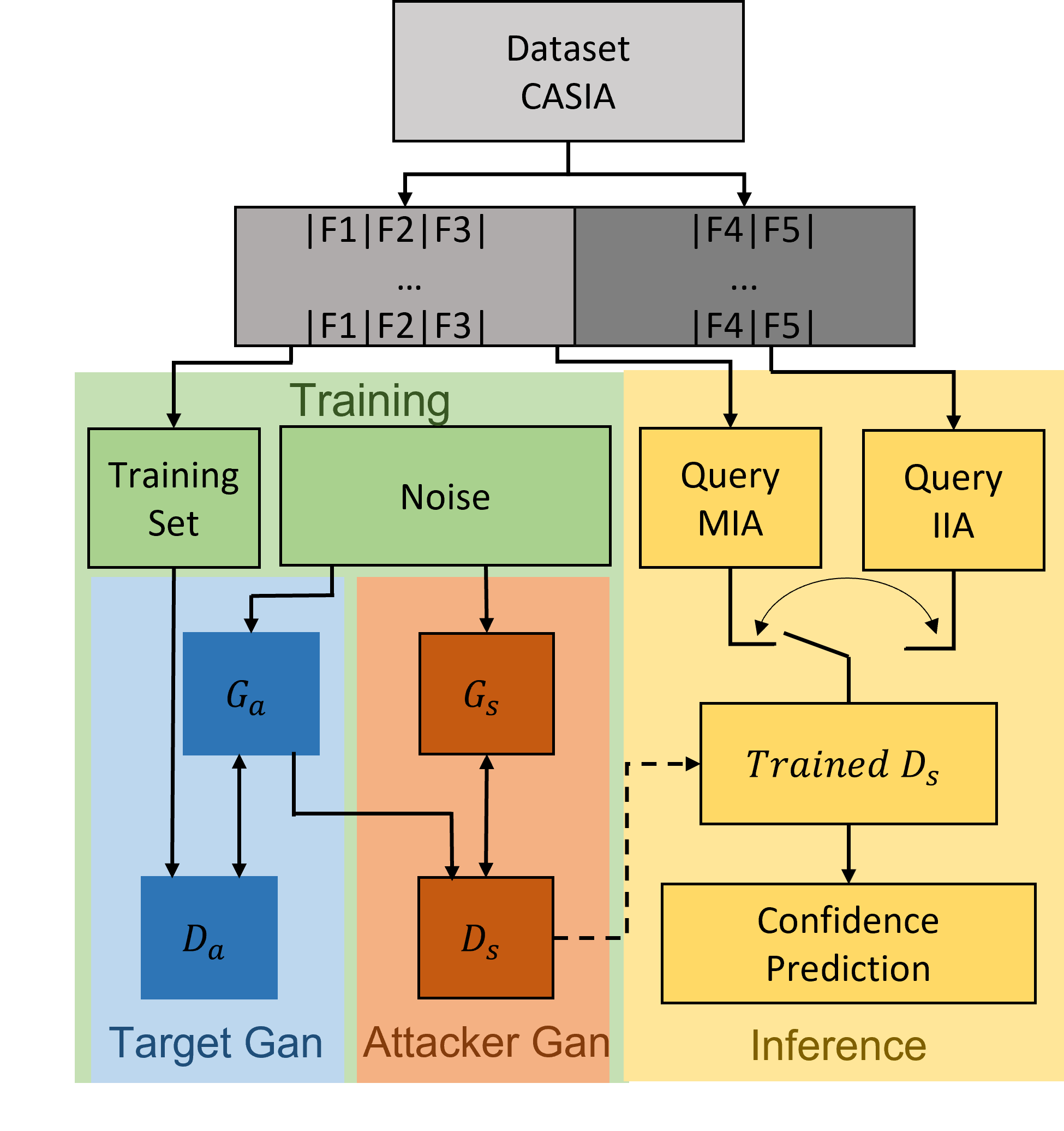}
    \caption{Schema of the attack}
    \vspace{-.5em}
    \label{fig:attack-scheme}
    \vspace{-.5em}
\end{figure}

The black box attack that we carry out against all the trained models is compliant with the state of the art concerning membership inference on generative networks proposed in in \cite{hayes2017logan} (see scheme in Figure~\ref{fig:attack-scheme}). Our main assumptions are that the attacker \sm{ has access to some APIs that allow generating from the target \gls{cgan} $ \{ \mathcal{G}_a, \mathcal{D}_a \} $  as many samples as he wants (the structure of the target model is not known) and he/she is in possession of a dataset containing some of the samples used to train  $ \{ \mathcal{G}_a, \mathcal{D}_a  \} $ } \smbis{(called the query dataset)}.

The attacker trains a shadow \gls{gan} $ \{ \mathcal{G}_s, \mathcal{D}_s  \} $ to mimic $ \{ \mathcal{G}_a, \mathcal{D}_a$ \}  . \smbis{More precisely, $\mathcal{G}_s$ is trained to generate samples as close as possible to those produced by $\mathcal{G}_a$, while the discriminator $\mathcal{D}_s$ classifies whether the sample was generated by $\mathcal{G}_a$ or $\mathcal{G}_s$}. In this way, $\mathcal{D}_s$ will \smbis{infer some peculiar features that $\mathcal{G}_a$ inadvertently introduces in the generated samples (likely due to overfitting)}.
At this point, \smbis{the confidence values outputted by $\mathcal{D}_s$ make it possible} to sort the samples in the query dataset. Ideally, the top $k$ samples are going to have a higher likelihood of being members \smbis{of the training set, thus achieving a successful \gls{mia}.} 

The main intuition behind this attack is that $\mathcal{D}_s$ should be able to better recognize images used to train $ \{ \mathcal{G}_a, \mathcal{D}_a  \} $ since they should present features that are more similar to the ones displayed in samples generated by $\mathcal{G}_a$.

In \cite{hayes2017logan} the authors show how the attack performance improves with the training time,
\sm{but knowing in advance a sufficient number of iterations permits reducing the computational burden.}

In this work, we show that by exploiting \gls{is} \cite{salimans2016improved} as the metric for early stopping it is possible to obtain good \gls{mia} performance. In particular, we stop training when the \gls{is} of the shadow model is similar to the one of the attacked one. 

\sm{

The score is proposed in \cite{salimans2016improved} as: 

\

\begin{equation}\exp(\mathbf{E}_x [KL(p(y|x)||p(y)]), \end{equation} where:

\begin{enumerate}
    \item KL is the \textbf{Kullback–Leibler divergence}, denoted by $D_{KL} ( P || Q )$, i.e., a statistical distance measuring how much the probability distribution P is different from a reference probability distribution Q.
    \item The results are modeled through exponentiation to allow easier comparisons between values.
\end{enumerate}}

\sm{The implementation of the Inception Score in the attacking strategy is motivated by the will of replicating a scenario where the attacker needs a quantitative method to assess the possible quality of his attempts. Leading the training of his networks according to the score obtained from the victims' generated images, the attacker increases his chances of replicating a truthful overfitting, hence creating shadow models able to replicate some characteristics of the victims' model behaviour.
Different metrics are often used in literature to quantify generated images quality, as the Frechet Inception Distance \cite{yu2021frechet}. Nevertheless, in our task based on black and white fingerprints, the IS implemented through a pretrained ImageNet classifier provided the necessary results. }

Lastly, we also explore the task of \gls{ulmia}. In this case, the aim is not only to determine if a given impression was used during training but also to infer if the specific finger (i.e., identity) itself was used in the training process. \sm{In this case the assumptions formulated above slightly change, in particular, the query dataset is modified so that it does not contain the same impression of the finger used in the training datasets. This is a more realistic scenario since it is unlikely for an attacker to use the same biometric samples.} The black box attack remains exactly identical to the one carried out in the \gls{mia} case since also in \gls{ulmia} we would like $\mathcal{D}_s$ to recognize features of the fingerprints that were present in the training set, and thus also on other impressions of the same fingerprints.

\section{Experimental Setup}
\label{sec:experimental}

\begin{figure}[!t]
    \centering
    \includegraphics[width=.165\textwidth]{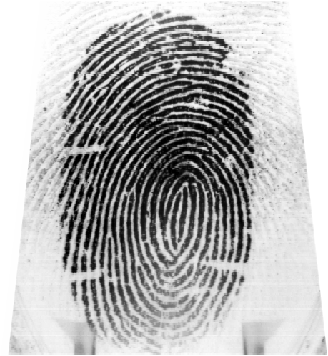}
    \includegraphics[width=.165\textwidth]{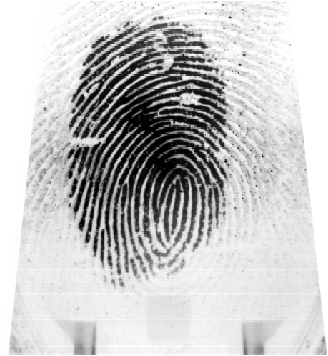}
    \includegraphics[width=.165\textwidth]{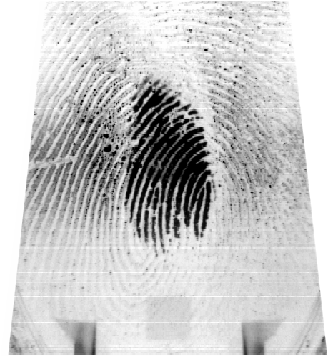}
    \includegraphics[width=.165\textwidth]{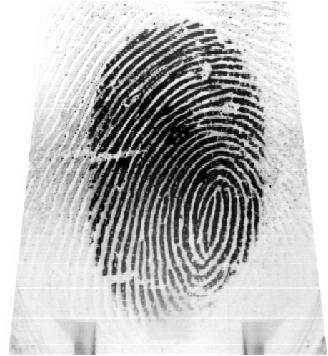}
    \vspace{-.5em}
    \caption{4 different impressions of the same finger}
    \vspace{-.5em}
    \label{fig:casia}
\end{figure}

\begin{figure}
    \centering
    \includegraphics[width=\columnwidth]{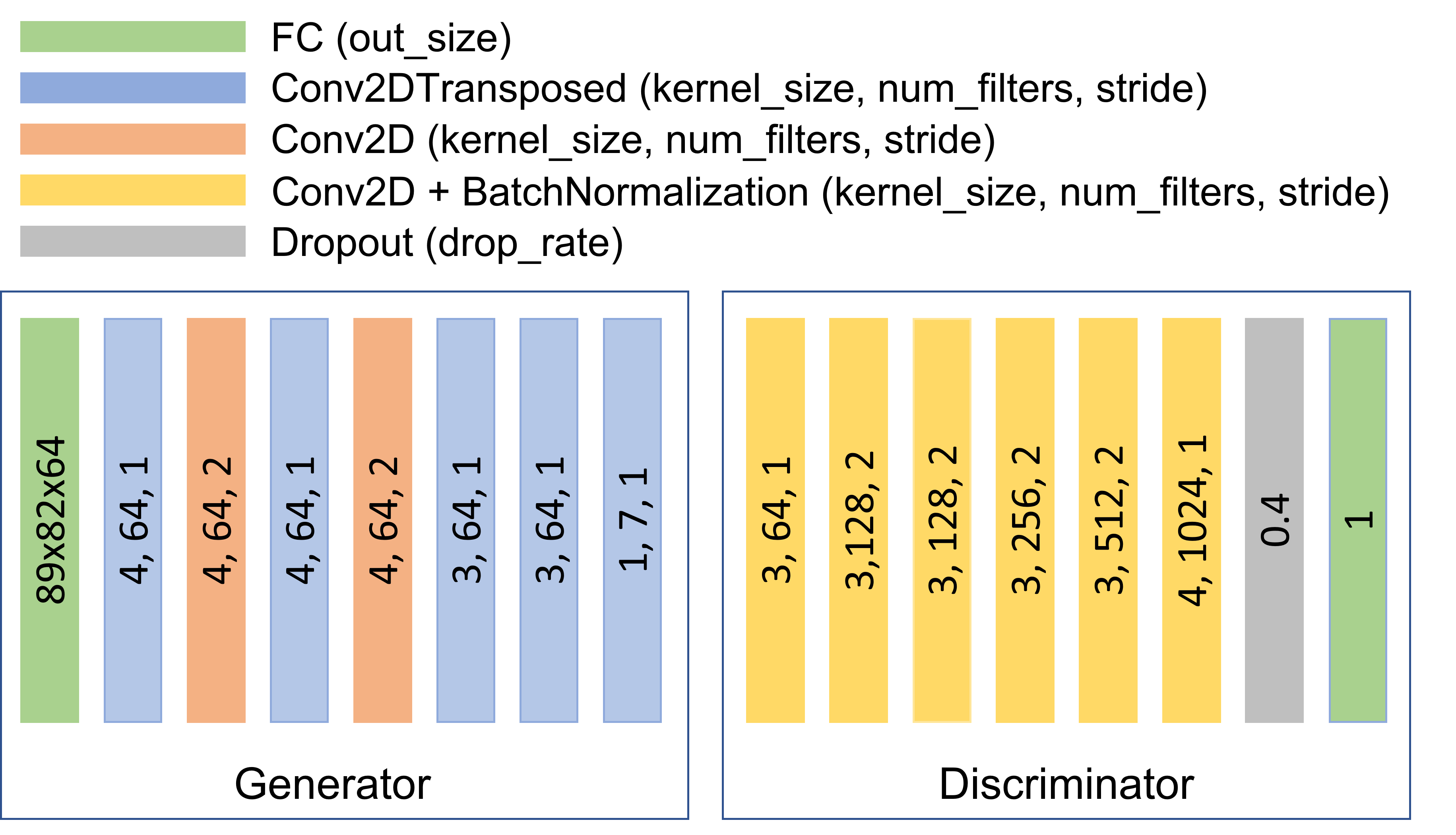}
    \vspace{-1em}
    \caption{Architectures of $\mathcal{G}_a, \mathcal{D}_a$}
    \label{fig:GAN-schema}
    \vspace{-1em}
\end{figure}

\sm{In our experimental set-up, we trained different \glspl{cgan} using impressions from the CASIA-FingerprintV5 dataset \cite{casia} which contains 20.000 fingerpint images of 500 subjects. These were captured using URU 4000 sensor. Each volunteer contributed with 40 fingerprint images obtained by getting five scans ($328\times 356$ resolution) out of eight different fingers. The volunteers varied rotation and pressure  generating significant differences in the quality of the different acquisitions (see Figure~\ref{fig:casia}).}

The attacked architecture is defined in Figure~\ref{fig:GAN-schema}, we use LeakyReLU as an activation function made exception for the last layer of the generator where we use TANH and in the last layer of the
discriminator where we use Sigmoid \smbis{(these implementation choices are rather generic and shared by most of the \gls{gan} architectures in the literature)}. The \smbis{attacking \gls{gan}} model was trained using Adam optimizer and the standard loss, however, we noticed that the discriminator was producing a very skewed prediction distribution making it hard to distinguish between samples where it had high and low confidence. To address this we added label smoothing with a smoothing factor equal to $0.2$.

From the attacker's point of view, in addition to the aforementioned change to the discriminator loss, following the requirements set by the Inception Score compatibility, all the main training parameters were modified (e.g., buffer size, batch size, epochs). $\mathcal{G}_s$ first layer was increased by half the size in the number of neurons, and the Adam optimizer was tuned accordingly. The attacking discriminators on the other hand was built adding convolutional layers.  Empirically, our results suggest that the quality of the inference (i.e., the relative overfitting) is jointly improved by both the widening of the $\mathcal{G}_s$ and the deepening and sharpening of $\mathcal{D}_s$ capability, with the latter introducing the highest influence (the attack produces decent results even with the generator being smaller and the discriminator being less deep, but without label smoothing the results are way less significant).

In order to assess the robustness of the trained models against \glspl{mia}, the considered \glspl{cgan} were retrained using all the dataset splits specified in Table \ref{tab:datasets}.  \sm{In all the data splits we only keep 3 acquisitions per finger in the training dataset, while the others are added to the \gls{ulmia} query dataset.} Table \ref{tab:datasets} is composed by the number of samples the victim's GAN used to train. The total number of samples is computed as $n\_fingers \times n\_user \times n\_fingerprints$, where $n\_fingers=8$, while $n\_user$ and $n\_fingerprints$ are specified in the table for each case.

\smbis{On the other hand, for what concerns the attacking model, we performed model selection using the \gls{is} metric (as mentioned above) similarly to what a black box attack would need to do in a real scenario.
As a result, we created diverse architectures with different parameters depending on the dataset that was originally used to train $ \{ \mathcal{G}_a, \mathcal{D}_a  \} $ . Also in this case, the different networks were obtained by stacking convolutional layers (with a structure similar to the one of the attacked \gls{gan}) whose number was varied in order to satisfy the \gls{is} criterion. The different configurations are here omitted for the sake of conciseness.}

\section{Results}
\label{sec:results}

\begin{figure}[t!]
    \centering
    \includegraphics[width=.165\textwidth]{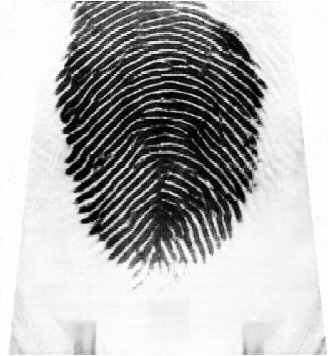}
    \includegraphics[width=.165\textwidth]{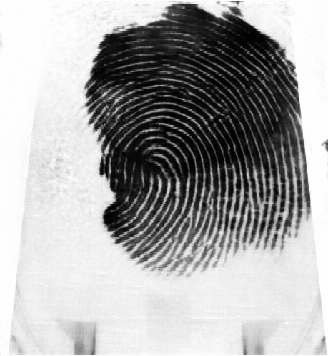}
    \includegraphics[width=.165\textwidth]{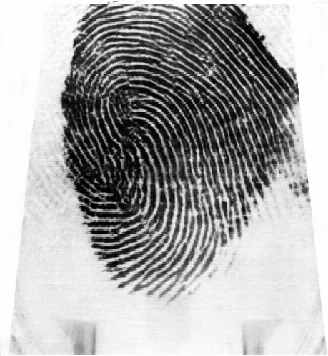}
    \includegraphics[width=.165\textwidth]{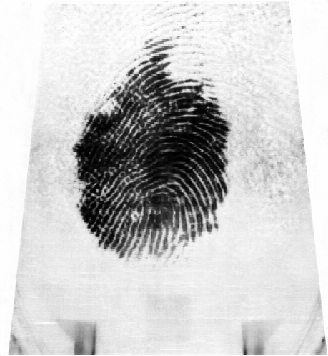}
    \vspace{-.5em}
    \caption{4 generated fingerprints impressions}
    \vspace{-.5em}
    \label{fig:generated}
\end{figure}

\begin{figure}[t!]
    \centering
    \includegraphics[width=.75\columnwidth]{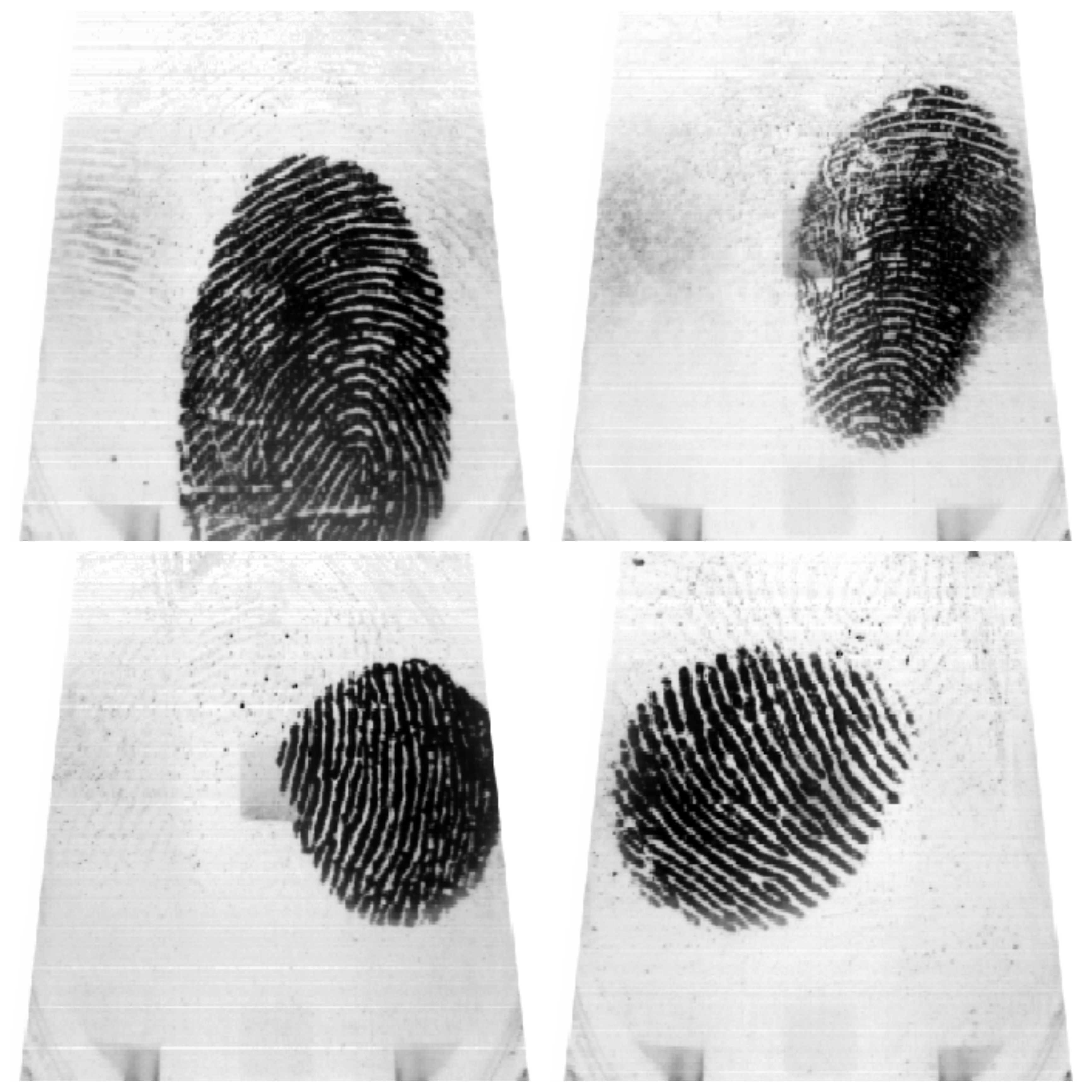}
    \vspace{-.5em}
    \caption{On the left, highest confidence fingerprints that are actual members. On the right highest scoring images that are different impressions of training samples. (Gan\_2400 and Gan\_4800)}
    \vspace{-1em}
    \label{fig:fingerprints}
\end{figure}

\begin{table*}[]
    \centering
    \begin{tabular}{|c|c|c|c|c|c|}
    \hline
     Name & Real Images & D1 & D2 & D3 & D4\\
    \hline
     Samples / Users / Fingerprints & 2000/50/5 & 9600/400/3 & 4800/200/3 &2400/100/3 & 1200/50/3 \\
     IS Mean & 1639.2 & 1620.13 & 1563.79 & 1570.89 & 1571.28 \\
     IS STD & 7.92 & 11.95 & 15.99 & 18.66 & 10.29 \\
    \hline
    \end{tabular}
    \vspace{-.5em}
    \caption{Datasets splits and inception score metrics on the trained GANs for each of them}
    \vspace{-.5em}
    \label{tab:datasets}
\end{table*}

\begin{table*}[t]
\begin{center}
\begin{tabular}{|l|c|c|c|c|c|c|}
\hline
& \multicolumn{3}{c|}{MIA} & \multicolumn{3}{c|}{IIA}\\
\hline
GAN & top 20 & top 200 & top 50\% & top 20 & top 200 & top 50\%\\
\hline
Gan\_9600 & 14/20 & 126/200 & 1172/2000 & 12/20 & 114/200 & 1161/2000\\
Gan\_4800 & 19/20 & 144/200 & 1133/2000 & 18/20 & 140/200 & 1133/2000\\
Gan\_2400 & 12/20 & 108/200 & 869/1600 & 13/20 & 105/200 & 869/1600 \\
Gan\_1200 & 13/20 & 132/200 & 526/800  & 12/20 & 137/200 & 540/800  \\

\hline
\end{tabular}
\end{center}
\vspace{-1em}
\caption{MIA and IIA Results.}
\vspace{-1em}
\label{tab:mia}
\end{table*}

\sm{
First experimental considerations concern the quality of the generated images and the effectiveness in using \gls{is} as a termination criterion.

Figure~\ref{fig:generated} shows that the generated fingerprint images consistently present level 1 (ridge orientation and singular points) and 2 (minutiae) features, with the sporadic presence of level 3 (pores) features as well. This is further highlighted by the \gls{is} values obtained on the generated images, which are comparable to those computed on fingerprints. It is also possible to notice that the lower the amount of training images the lower the \gls{is} value.}

We implement the statistical attack on a query dataset built with two sets of images: 
\begin{enumerate}
    \item samples associated to users that did not contribute to the attacked gan ($G_a$) training set.
    \item samples associated to users that contributed to the attacked gan ($G_a$) training set.
\end{enumerate}
The former are the negative samples, i.e., the ones that should not be detected as being part of the dataset, while the latter are the positive samples. For MIA the positive samples are those that were actually used to train the attacked gan $G_a$ while in the case of IIA they are other impressions of fingerprints used to train $G_a$.

Since four GANs were trained with a decreasing number of samples (i.e., fewer users contributed to the dataset), to assess the effect of the dataset size on the attack performance, the query datasets can contain at most as many fingerprints as the ones available for the IIA attacks (i.e., $2number\_training\_users$). For this reason two query datasets contain 4000 samples (2000 positive and 2000 negative)  while two have fewer fingerprints, respectively 3200 (1600 positive and 1600 negative) and 1600 (800 positive and 800 negative) samples, since fewer users were used to train the GAN.

Following previous works from the literature we report the performance of the algorithm in terms of positive samples in the top-20, top-200 and top-50\% samples (here we use top-50\% instead of top-2000 because, as aforementioned, in two cases fewer than 4000 query samples were available) sorted according to the rank assigned to the fingerprints by the shadow discriminator. Note that the top-50\% configuration should have been top-2000,  however, since 2000 samples per query dataset only the top 50\% samples were considered)
In such a scenario, if the discriminator was actually random guessing, the number of positive and negative samples would be very close to 10, 100 and 25\% of the total for all the models. However, although this happens at time, in all cases reported in Table \ref{tab:mia} the number of positive samples is greater than the number of negative samples at times being very close to 100\% (18/20, 19/20, etc.) showing that there is indeed some sort of information leakage.

Additionally, it is possible to see that the performance of the attack is only slightly worse when performing \gls{ulmia} compared to \gls{mia} \smbis{showing that, even in the more challenging scenario, private information about the users that contributed in the dataset creation is leaked to the attacker.}

This demonstrates the effectiveness of \gls{mia} and \gls{ulmia} in detecting sample and finger membership in the training dataset for a \gls{gan}. As a matter of fact, the attacks lead to near-perfect recognition of many fingerprints as belonging to the targeted sample in both scenarios.

\smbis{An additional proof that the success of the attack is provided by a visual inspection of images where $\mathcal{D}_s$ confidence suggests that they are more likely to belong to the training dataset. These indeed exhibit various similarities especially within the Level 1 features as can be seen from Figure~\ref{fig:fingerprints}.}

Especially in the case of the \gls{gan} trained on 4800 samples, out of the 20 flagged images with the highest scores, 19 of them are indeed from the training dataset. Once the actual training images are swapped with a set of "remaining" acquisitions, i.e., different samples of the same fingers, the result remains high with 18 out of 20 correctly identified showing how much information these types of models can actually leak even though they were designed in the first place to avoid privacy concerns.

\section{Discussion}
\label{sec:discussion}
From Table~\ref{tab:mia} it is possible to see that all the trained models could be attacked by the proposed approach. \smbis{Additionally, since the attack was successful on all the four cases, it is possible to conclude that  \gls{is} might be a good heuristic when choosing when to stop the training procedure.} 

\smbis{Despite this, experimental tests did not reveal how the number of training samples affects the attack performance since the general success is also strongly dependent on other factors such as how well-trained the attacked model is or the batch size and optimizer, to mention some of them.}
\smbis{However, considering other results on \gls{mia} presented in the literature, it is reasonable to believe that increasing the total number of samples by a significant amount implies improving the robustness against \glspl{mia} and \glspl{ulmia}.}

\section{Conclusion}
\label{sec:conclusion}
\smbis{This work illustrates the main vulnerabilities of generative adversarial networks as a mean to solve the data shortage and privacy issues for learned biometric architectures. More precisely, Membership and Identity Inference Attacks on fingerprint \glspl{gan} are described and evaluated showing how it is possible to infer users' identity from a trained black box model.} 

\dm{To the best of our knowledge, we are the first to notice the severity of the problem on fingerprint data showing that it is possible to assess membership of a sample and to detect if a person contributed to the creation of a dataset without having access to any of the training data.

A possible follow up implementation of this work stands within the specific-sample research methods. Here we proved the vulnerability on the information leakage of the discussed models. By looking at the distributions of the shadow discriminator, it is possible for the attacker to notice significant patterns that can help to predict whether new fingerprints are related to the training set of the attacked model or not. In Figure \ref{fig:kernels} we show plots of distribution kernels for the attacking queries in two instances (1200, 4800). We propose to use the information given by these distributions to create classifications methods able to cluster unforeseen data to allow the attackers to link it to the training set. A possible approach could be the training of either supervised or unsupervised models to respectively learn to fit the relative distributions or to learn intrinsic patterns that differentiate the shadow discriminator judgement. 

\sm{Being the proposed scheme quite generalizable, our future research will analyze IIA problem on other kinds of biometric samples and architectures, and will focus on the design of defense strategies.}
\begin{figure}[t!]
    \centering
    \includegraphics[width=.235\textwidth]{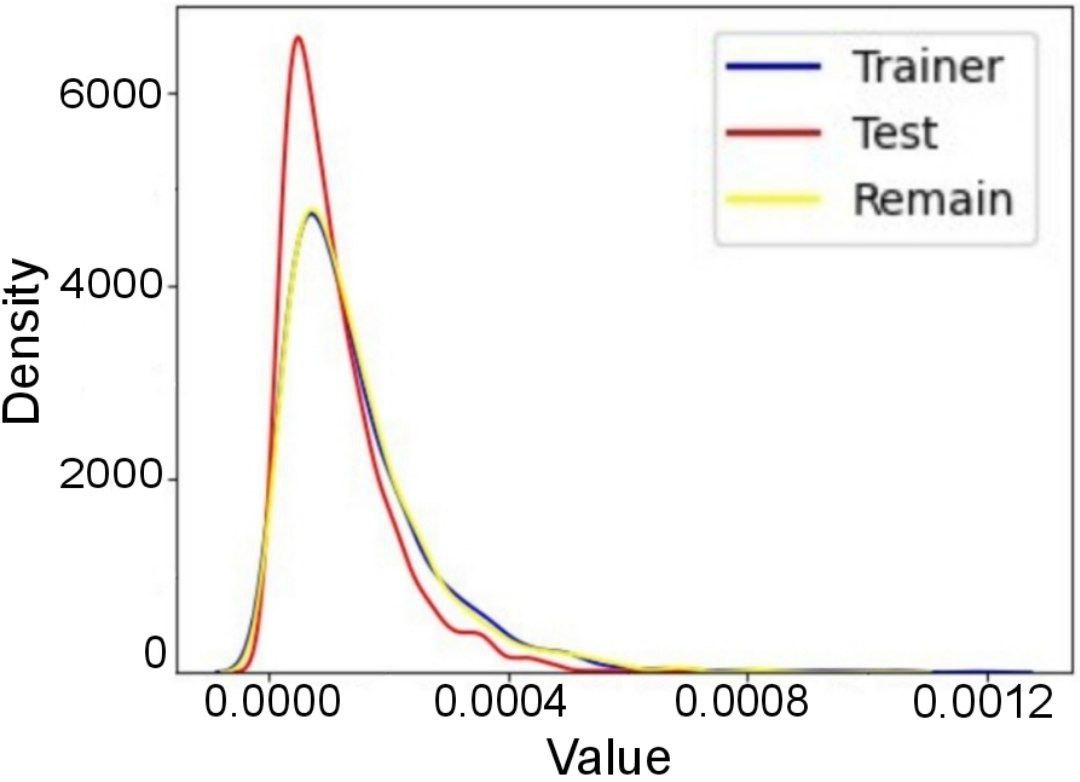}
    \includegraphics[width=.225\textwidth]{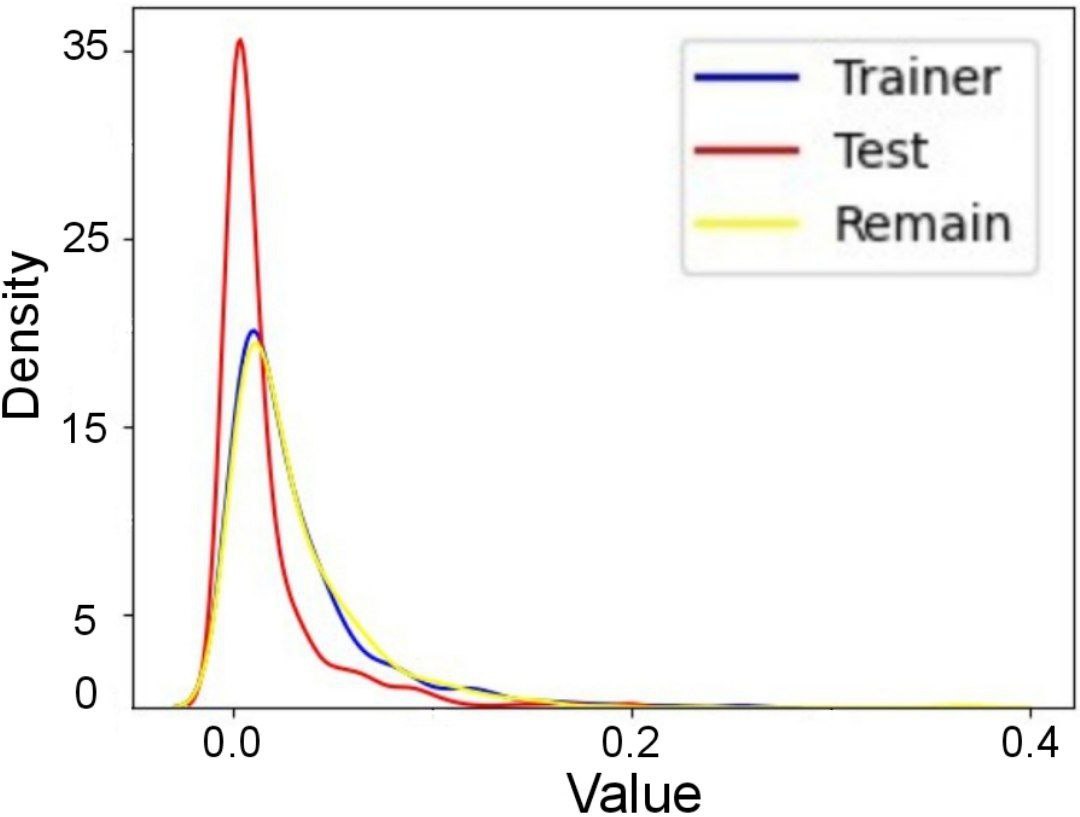}
    \vspace{-.8em}
    \caption{Kernels distributions of the attacking discriminators against the generated images of the Gans with 1200 and 4800 samples.}
    \vspace{-.8em}
    \label{fig:kernels}
\end{figure}
}

\bibliographystyle{elsarticle-num} 
\bibliography{biblio.bib}
\end{document}